\documentclass[10pt,twocolumn,letterpaper]{article}

\usepackage{cvpr}
\usepackage{times}
\usepackage{epsfig}
\usepackage{graphicx}
\usepackage{amsmath}
\usepackage{amssymb}
\usepackage{multirow}
\usepackage{mathrsfs}
\usepackage{booktabs}
\usepackage[T1]{fontenc}
\usepackage[utf8]{inputenc}
\usepackage{authblk}

\usepackage[breaklinks=true,bookmarks=false]{hyperref}

\cvprfinalcopy 

\def\cvprPaperID{1423} 

\begin{document}

\title{Attentive Relational Networks for Mapping Images to Scene Graphs}

\author[1,2]{Mengshi Qi$^*$}
\author[3]{Weijian Li\thanks{Equal contribution.}}
\author[3]{Zhengyuan Yang}
\author[1,2]{Yunhong Wang\thanks{Corresponding author.}}
\author[3]{Jiebo Luo$^\dag$}
\affil[1]{State Key Laboratory of Virtual Reality Technology and Systems\authorcr School of Computer Science and Engineering, Beihang University, China}
\affil[2]{Beijing Advanced Innovation Center for Big Data and Brain Computing}
\affil[3]{Department of Computer Science, University of Rochester, USA}
\affil[ ]{\textit {\{qi\underline{\hbox to 2mm{}}mengshi, yhwang\}@buaa.edu.cn, \{wli69,zyang39,jluo\}@cs.rochester.edu}}

\maketitle

\pagestyle{empty}  
\thispagestyle{empty} 

\begin{abstract}
	Scene graph generation refers to the task of automatically mapping an image into a semantic structural graph, which requires correctly labeling each extracted object and their interaction relationships. Despite the recent success in object detection using deep learning techniques, inferring complex contextual relationships and structured graph representations from visual data remains a challenging topic. In this study, we propose a novel Attentive Relational Network that consists of two key modules with an object detection backbone to approach this problem. The first module is a semantic transformation module utilized to capture semantic embedded relation features, by translating visual features and linguistic features into a common semantic space. The other module is a graph self-attention module introduced to embed a joint graph representation through assigning various importance weights to neighboring nodes. Finally, accurate scene graphs are produced by the relation inference module to recognize all entities and the corresponding relations. We evaluate our proposed method on the widely-adopted \emph{Visual Genome Dataset}, and the results demonstrate the effectiveness and superiority of our model.  
\end{abstract}


\section{Introduction}
\label{sec:intro}

\begin{figure}[t]
	\centering
	\resizebox{\columnwidth}{!}{\includegraphics{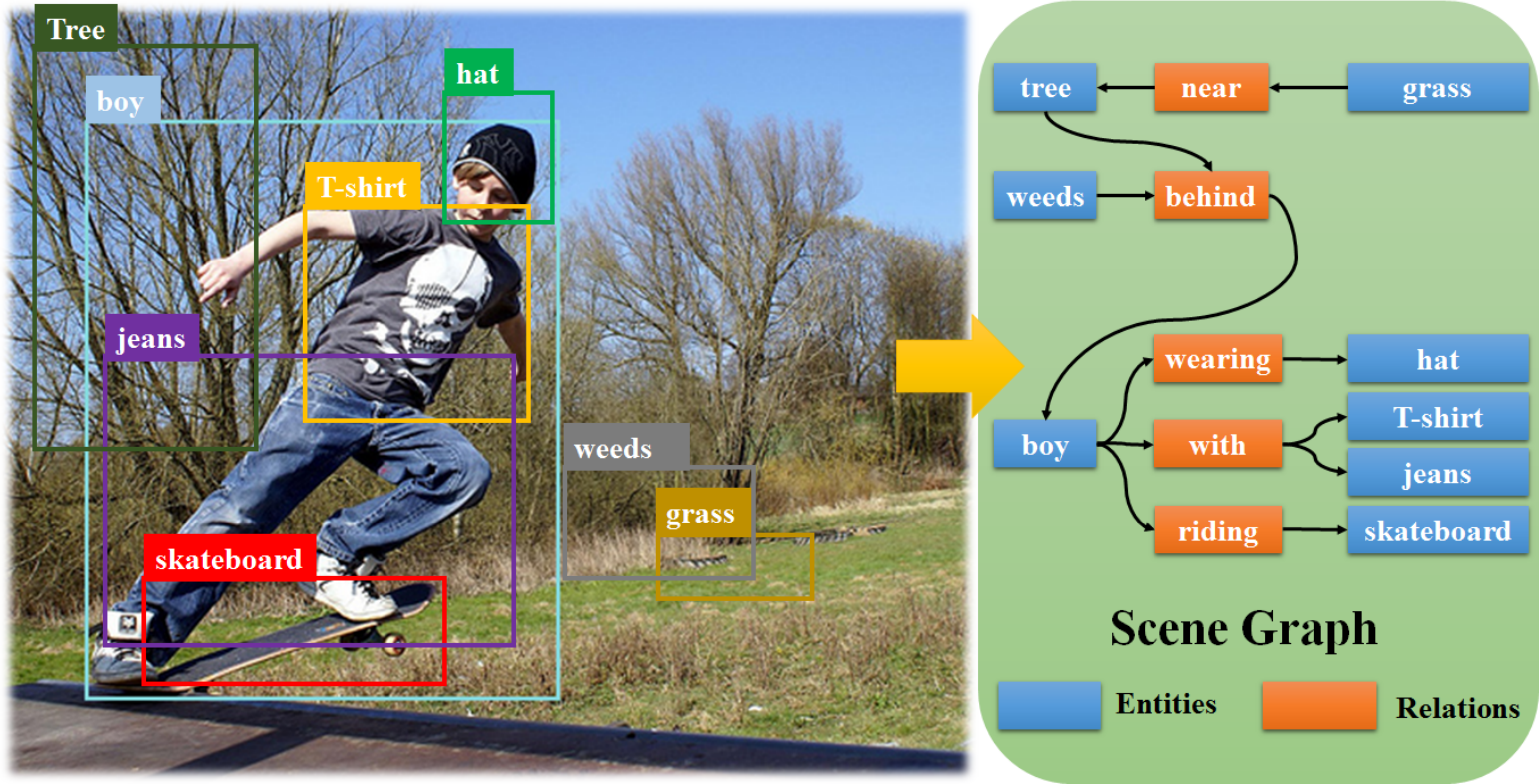}}
	\vspace{-3mm}
	\caption{Illustration of the task of scene graph generation. Using our proposed Attentive Relational Network, an image can be mapped to a scene graph, which captures individual entities~(\eg 
		boy, tree and grass) and their relationships~(\eg $<$boy-riding-skateboard$>$ and $<$weeds-behind-boy$>$).}
	\label{fig:issue}
	\vspace{-5mm}
\end{figure}

Visual scene understanding~\cite{johnson2015image,lazebnik2006beyond,zhou2014learning} is a fundamental problem in computer vision. It aims at capturing the structural information in an image including the object entities and pair-wise relationships. As is shown in Figure~\ref{fig:issue}, each entity and relation should be processed with a broader context to correctly understand the image at the semantic level. During recent years, deep neural network based object detection models such as Faster R-CNN~\cite{girshick2015fast,ren2017faster} and YOLO~\cite{redmon2016you} have achieved great improvements. However, such conventional object detection approaches cannot capture and infer the relationships within an image.

Because of its ability to enrich semantic analysis and clearly describe how objects interact with each other~(\eg ``a boy is riding a skateboard'' in Figure~\ref{fig:issue}), generating scene graphs from images plays a significant role in multiple computer vision applications, such as image retrieval~\cite{johnson2015image,qi2017online}, image captioning~\cite{li2019know,li2017scene,yao2018exploring}, visual question answering~\cite{li2017vip,teney2017graph} and video analysis~\cite{qi2018stagnet,yuan2017temporal}. The highly diverse visual appearances and the large numbers of distinct visual relations make scene graph generation a challenging task.

Previous scene graph generation methods~\cite{herzig2018mapping,li2018factorizable,li2017scene,newell2017pixels,xu2017scene,yang2018graph,zellers2018neural} locate and infer the visual relationship as a triplet in the form $<$subject-predicate-object$>$, and the predicate is a word used to link a pair of objects, \eg $<$boy-wearing-hat$>$ in Figure~\ref{fig:issue}. There exist various kinds of relationships between two objects, including spatial positions~(\eg under, above), attributes/ prepositions~(\eg with, of), comparatives~(\eg taller, shorter) and actions/ verb~(\eg play, ride). Most of the existing works neglect the semantic relationship between the visual features and linguistic knowledge, and the intra-triplet connections.

Moreover, previous works invariably utilize conventional deep learning models such as Convolutional Neural Networks~(CNN)~\cite{li2018factorizable,li2017scene,newell2017pixels,yang2018graph} or Recurrent Neural Networks~(RNN)~\cite{herzig2018mapping,xu2017scene,zellers2018neural} for scene graph generation. These methods require to know the graph structure beforehand and contain computationally intensive matrix operations during approximation. In addition, most of them follow a step-by-step manner to capture the representation of nodes and edges, leading to neglect the global structure and information in whole image. Effectively extracting a whole joint graph representation to model the entire scene graph for reasoning is promising but remains an arduous problem.

To address the aforementioned issues, we propose a novel \emph{Attentive Relational Network} that maps images to scene graphs. To be specific, the proposed method first adopts an object detection module to extract the location and category probability of each entity and relation. Then a semantic transformation module is introduced to translate entities and relation features as well as their linguistic representation into a common semantic space. In addition, we present a graph self-attention module to jointly embed an adaptive graph representation through measuring the importance of the relationship between neighboring nodes. Finally, a relation inference module is leveraged to classify each entity and relation by a Multi-Layer Perceptron~(MLP), and to generate an accurate scene graph. Our main contributions are summarized as follows:
\begin{itemize}
	\item A novel Attentive Relational Network is proposed for scene graph generation, which translates visual information to a graph-structured representation.  
	\item A semantic transformation module is designed to incorporate relation features with entity features and linguistic knowledge, by simultaneously mapping word embeddings and visual features into a common space.
	\item A graph self-attention module is introduced to embed the joint graph representation by implicitly specifying different weights to different neighboring nodes. 
	\item Extensive experiments on the \emph{Visual Genome Dataset} verify the superior performance of the proposed method compared to the state-of-the-art methods. 
\end{itemize}


\section{Related Work}
\label{sec:rel}

{\bf Scene Graph Generation.}~Significant efforts have been devoted to this task during recent years, which can be divided into two categories: Recurrent Neural Networks~(RNN)-based methods~\cite{herzig2018mapping,xu2017scene,zellers2018neural} and Convolutional Neural Networks~(CNN)-based approaches~\cite{li2018factorizable,li2017scene,newell2017pixels,yang2018graph}. Xu~\emph{et al.}~\cite{xu2017scene} employ RNNs to infer scene graphs by message passing. Zellers~\emph{et al.}~\cite{zellers2018neural} introduce~\emph{motifs} to capture the common substructures in scene graphs. To minimize the effect of different input factors' order, Herzig~\emph{et al.}~\cite{herzig2018mapping} propose a permutation invariant structure prediction model. Li~\emph{et al.}~\cite{li2017scene} construct a dynamic graph to address multi tasks jointly. While Newell~\emph{et al.}~\cite{newell2017pixels} present an associative embedding technique~\cite{newell2017associative} for predicting graphs from pixels. Yang~\emph{et al.}~\cite{yang2018graph} propose a Graph R-CNN by utilizing graph convolutional network~\cite{kipf2016semi} for structure embedding. Li~\emph{et al.}~\cite{li2018factorizable} present a Factorizable Net to capture subgraph-based representations. Unlike previous work, our proposed model focuses on discovering semantic relations through jointly embedding linguistic knowledge and visual representations simultaneously.

\begin{figure*}[htbp]
	\centering
	\includegraphics[width=0.95\textwidth]{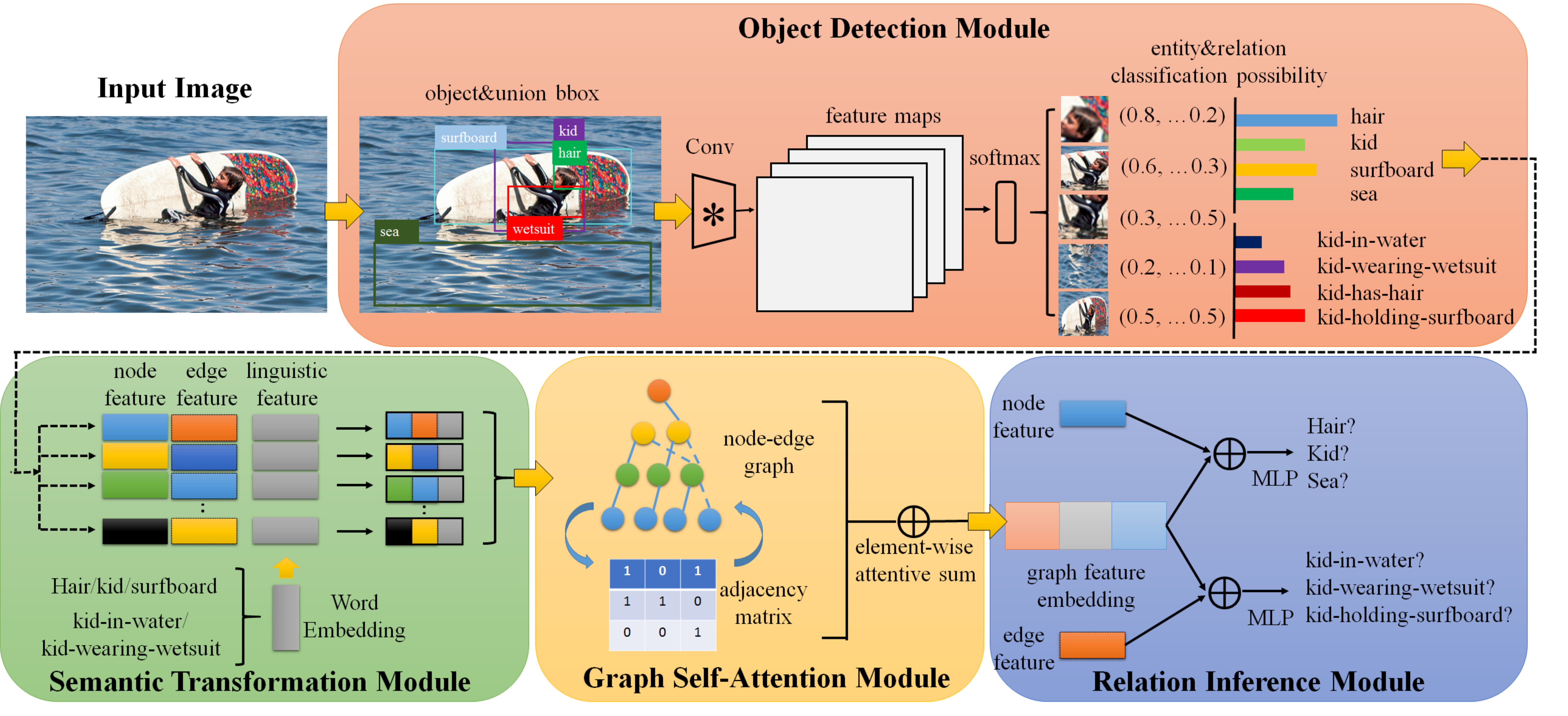}
	\caption{Overview of the proposed Attentive Relational Network. Our model mainly consists of four parts:~(1)~{\it Object Detection Module}:~capturing the visual feature and the location of each entity bounding box with their pair-wise relation bounding boxes. Then a softmax function is employed to obtain initial classification scores for each entity and relation; (2)~{\it Semantic Transformation Module}: producing the semantic embedded representations by transforming label word embeddings and visual features into a common semantic space; (3)~{\it Graph Self-Attention Module}: leveraging a self-attention mechanism to embed entities via constructing an adjacency matrix based on the space position of nodes; (4)~{\it Relation Inference Module}: creating the joint global graph representation and predicting entity and relationship labels as final scene graph result.}
	\label{fig:framework}
	\vspace{-5mm}
\end{figure*}

{\bf Visual Relationship Detection.}~Early efforts in visual relationship detection ~\cite{atzmon2016learning,farhadi2010every,ramanathan2015learning,sadeghi2011recognition} tend to adopt a joint model regarding the relation triplet as a unique class. The visual embedding-based approaches~\cite{lu2016visual,woo2018linknet,yu2017visual,zhang2017visual,zhuang2017towards} place objects in a low-dimensional relation space and integrates extra knowledge. However, these works can not learn graph structural representation, which denotes the positional and logical relationships between objects in the image. Plummer~\emph{et al.}~\cite{plummer2017phrase} combine different cues with learning weights for grounded phrase. Liang~\emph{et al.}~\cite{liang2017deep} adopt variation-structured reinforcement learning to sequentially discover object relationships. Dai~\emph{et al.}.~\cite{dai2017detecting} exploit the statistical dependencies between objects and their relationships. Recently, various studies~\cite{hwangtensorize,krishna2018referring,li2017vip,peyre2017weakly,yang2018shuffle,yin2018zoom,zhang2017ppr,zhang2018grounding,zhang2017relationship} propose relationship proposal networks by employing pair-wise regions for fully or weakly supervised visual relation detection. However, most of them are designed for detecting relationship one-by-one, which is inappropriate for describing the structure of the whole scene. Our proposed graph self-attention based model aims at embedding a joint graph representation to describe all relationships, and applying it for scene graph generation. 

\section{Proposed Approach} 
\label{sec:app}

\subsection{Overview}
\label{subsec:over}

\textbf{Problem Definition:}~We define the \emph{scene graph} of an image $I$ as $G$, which describes the category of each entity and semantic inter-object relationships. A set of entity bounding boxes as $B=\{b_1, ..., b_n\}, b_i\in\mathbb{R}^4$ and their corresponding class label set $O=\{o_1, ..., o_n\}, o_i\in C$, where $C$ is object categories set. The set of binary relationships between objects are referred to as $R=\{r_1, ..., r_m\}$. Each relationship $r_k\in R$ is a triplet in a $<$subject-predictive-object$>$ format, where a subject node $(b_i, o_i)\in B\times O$, a relationship label $l_{ij}\in\mathcal{R}$ and an object node $(b_j, o_j)\in B\times O$. $\mathcal{R}$ is the set of all predicates\footnote{We also adding extra `bg' referred to `background', denoting there is no relationship or edge between objects.}.

{\bf Graph Inference:}~Each Scene graph comprises of a collection of bounding boxes $B$, entity labels $O$ and relation labels $R$. The possibility of inferring a scene graph from an image can be formulated as the following:
\begin{equation}
	Pr(G|I)=Pr(B|I)Pr(O|B,I)Pr(R|B,O,I).
\end{equation}
The formulation can be regarded as the factorization without independence assumptions. $Pr(B|I)$ can be inferred by the object detection module in our model described in~\ref{subsec:object}, while $Pr(O|B,I)$ and $Pr(R|B,O,I)$ can be inferred by the rest of modules proposed in our model.

Figure~\ref{fig:framework} presents the overview of our proposed Attentive Relational Network, which contains four modules, namely object detection module, semantic transformation module, graph self-attention module and relation inference module. Our model aims at producing a joint global graph representation for the image, which contains the semantic relation translated representation learned in semantic transformation module, and the whole entity embedded representation captured in graph self-attention module. Finally, we combine the learned global graph representation and each entity/relation feature for reasoning in relation inference module. Next we will respectively introduce the four proposed modules in detail. 

\subsection{Object Detection Module}
\label{subsec:object}

We employ Faster R-CNN~\cite{ren2017faster} as our object detector. Then a set of predictable entity proposals $B=\{b_1, ..., b_n\}$ from each input image $I$, including their locations and appearance features, are obtained. In order to represent the contextual information for visual relation, we generate an union bounding box to cover object pairs with a small margin. Two types of features can be adopted for describing entities and relations, \ie the appearance feature and the spatial feature~(the coordinates of the bounding box). Finally, we utilize the softmax function to recognize the category of each entity and relation, and achieve their corresponding classification confidence scores as the initial input to the following modules.

\subsection{Semantic Transformation Module}
\label{subsec:seman}

\begin{figure}[t]
	\centering
	\resizebox{8.5cm}{!}{\includegraphics{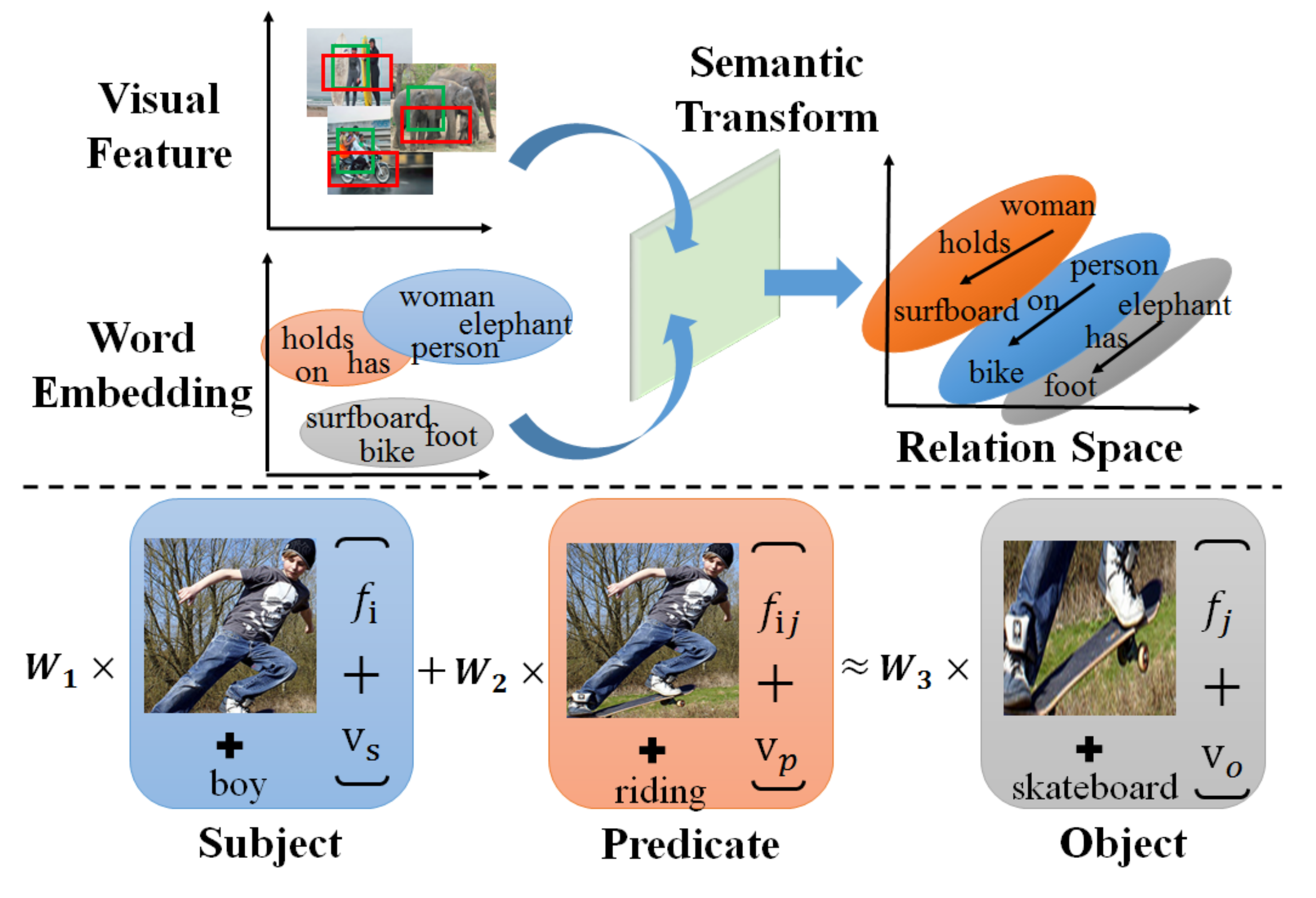}}
	\vspace{-3mm}
	\caption{Illustration of Semantic Transformation Module.~(Top):~Mapping visual feature and word embedding to a common semantic space, and inferring their relationship in the relation space.~(Bottom): An example of relation translation. Concatenating the visual features of entities and relation~(\ie $f_i$, $f_j$ and $f_{ij}$) and their corresponding label embedding features~(\ie `boy', `riding' and `skateboard':~$v_s$, $v_p$ and $v_o$), and translating them based on $<$subject-predicate-object$>$ template via learned weight matrices~(\ie $W_1$, $W_2$ and $W_3$). }
	\label{fig:semantic}
	\vspace{-3mm}
\end{figure}
Inspired by Translation Embedding~(TransE)~\cite{bordes2013translating,zhang2017visual} and visual-semantic embedding~\cite{frome2013devise}, we introduce a semantic transformation module to effectively represent $<$subject-predicate-object$>$ in the semantic domain. As depicted in Figure~\ref{fig:semantic}, the proposed module leverages both visual features and textual word features to learn the semantic relationship between pair-wise entities. It then explicitly maps them into a common relation space. For any relation, we define $v_s$, $v_p$ and $v_o$ to represent the word embedding vectors of category labels for \emph{subject}, \emph{predicate} and \emph{object}. To generate specific word embedding vectors for subject, predicate and object, label scores obtained from Object Detection Module and word embedding of all labels are combined with element-wise multiplication. In computational linguistics, it is known that a valid semantic relation can be expressed as the following~\cite{pennington2014glove}: 
\begin{equation}
	\label{eq:trans_v}
	v_s+v_p\approx v_o,
\end{equation}
Similarly, we assume such a semantic relation exists among the corresponding visual features:
\begin{equation}
	\label{eq:trans_f}
	f_i+f_{ij}\approx f_j,
\end{equation}
where $f_i$, $f_j$ and $f_{ij}$ are defined as the visual representations of entity $b_i$, $b_j$ and their relation $r_{ij}$, respectively.

It is worth noting that the visual feature and word embedding should be projected into a common space. Hence, we adopt a linear model with three learnable weights to jointly approximate Eq.~(\ref{eq:trans_v}) and Eq.~(\ref{eq:trans_f}). L2 loss is used to guide the learning process:
\begin{equation}
	\mathcal{L}_{semantic}=\|W_3\cdot [f_j,v_o]-(W_1\cdot [f_i,v_s]+W_2\cdot [f_{ij},v_p])\|^2_2,
\end{equation}
where $W_1$, $W_2$ and $W_3$ refer to the weights respectively, and $[\cdot]$ denotes the concatenation operation. These learned weight matrices can be regarded as the semantic knowledge in relation space.

Then we need to map the visual features of detected entities (\ie nodes) and relations (\ie edges) with such linguistic knowledge into a common semantic domain. The semantic transformed representation of relation $f_{ij}$ in the scene graph can be denoted as $\Theta(f_{ij})$:
\begin{equation}
	\label{eq:relation}
	\Theta(f_{ij})=[(W_1\cdot [f_i,v_s]), (W_2\cdot [f_{ij},v_p]), (W_3\cdot [f_j,v_o])],
\end{equation}
where $[\cdot]$ denotes concatenation operation. Then we obtain the embedded representation of each relation in an image. 

\subsection{Graph Self-Attention Module}
\label{subsec:graph}

\begin{figure}[t]
	\centering
	\resizebox{7.0cm}{!}{\includegraphics{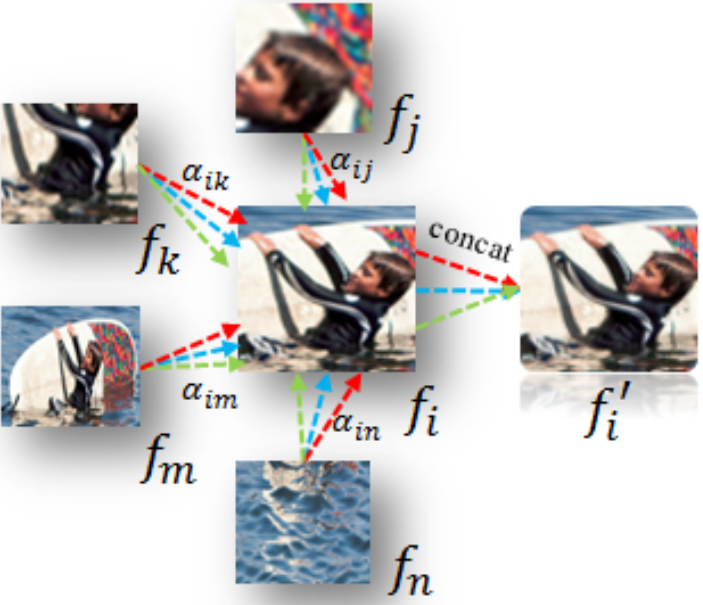}}
	\vspace{-1.3mm}
	\caption{Illustration of Graph Self-Attention Module for each single node. The output feature of the $i$-th node can be calculated based on its neighboring nodes' features $f_j$, $f_k$, $f_m$ and $f_n$ with their corresponding pair-wise attention weight $\alpha$. Different color arrows refer to independent attention computations as multi-head attention~(\eg k=3 in this figure). The aggregated attentive feature of node $i$ is denoted as $f^{'}_i$ via concatenation operation.}
	\label{fig:attention}
	\vspace{-3mm}
\end{figure}
The attention mechanism maps the input to a weighted representation over the values. Especially, self-attention has been demonstrated to be effective in computing representations of a single sequence~\cite{kipf2016semi,vaswani2017attention,velickovic2017graph}. To compute a relational representation of a singe node sequence, we introduce a graph self-attention module that takes both node representations and their neighborhood features into consideration. By adopting the self-attention mechanism, each node's hidden state can be extracted by attending over its neighbors and simultaneously preserve the structural relationship. 

As shown in Figure~\ref{fig:attention}, we define a collection of input node~(entity) features $F_{node}=\{f_1, f_2, ..., f_N\}, f_i\in\mathbb{R}^M$, and their corresponding output features $F'_{node}=\{f'_1, f'_2, ..., f'_N\}, f'_i\in\mathbb{R}^{M'}$, where $N$, $M$ and $M'$ are the number of nodes, input feature dimension and output feature dimension respectively. The attention coefficients $e_{ij}$ can be learned to denote the importance of node $j$ to node $i$:
\begin{equation}
	e_{ij}=\Lambda(U\cdot f_i, U\cdot f_j),
\end{equation}
where $\Lambda$ denotes attention weight vector implemented with a single feed-forward layer. $U\in\mathbb{R}^{M'\times M}$ refers to learnable parameter weight. 

We compute the $e_{ij}$ for each neighboring node $j\in\mathbb{N}_i$, where $\mathbb{N}_i$ denotes the neighboring set of node $i$. Then we normalize the coefficients across all neighboring nodes by the softmax function, for effective comparison with different nodes:
\begin{equation}
	\alpha_{ij}=\textrm{softmax}_{j}(e_{ij})=\frac{\exp(e_{ij})}{\sum_{k\in \mathbb{N}_i}\exp(e_{ik})}.
\end{equation}
Therefore the coefficients computed can be formulated as:
\begin{equation}
	\alpha_{ij}=\frac{\exp(\phi(\Lambda^T[U\cdot f_i, U\cdot f_j]))}{\sum_{k\in \mathbb{N}_i}\exp(\phi(\Lambda^T[U\cdot f_i, U\cdot f_k]))},  
\end{equation}
where $\phi$ and $[\cdot]$ represent Leaky ReLU nonlinear activation and concatenation operation. Final node representation is then obtained by applying the attention weights on all the neighboring node features. Inspired by~\cite{vaswani2017attention}, we employ multi-head attention to capture different aspect relationships from neighboring nodes. The overall output of the $i$-th node is a concatenated feature through $K$ independent attention heads, denoted as $\Phi(f_i)$: 
\begin{equation}
	\label{eq:multihead}
	\Phi(f_i)=\textrm{Concat}^K_{k=1}\sigma(\sum_{j\in \mathbb{N}_i}\alpha^k_{ij}U^kf_j),
\end{equation}
where $\alpha^k_{ij}$ are normalized attention coefficients by the $k$-th attention mechanism, $\sigma$ refers to nonlinear function, and $U^k$ is the input linear transformation's weight matrix\footnote{In our experiments, we set k=8 following~\cite{vaswani2017attention}.}.

{\bf Setting of Adjacent Matrix:}~In order to compute adjacent matrices, we design four strategies to determine node neighbors based on spacial clues. Concretely, given two bounding boxes $b_i$ and $b_j$ as two nodes, their normalized coordinates of locations can be denoted as $(x_i, y_i)$ and $(x_j,y_j)$, and their distance can be denoted as $d_{ij}=\sqrt{(x_j-x_i)^2+(y_j-y_i)^2}$. Then four neighbor classification settings are:~(1)~Inside Neighbor:~if $b_i$ completely includes $b_j$; (2)~Cover Neighbor:~if $b_i$ is fully covered by $b_j$; (3)~Overlap Neighbor:~if the IoU between $b_i$ and $b_j$ is larger than 0.5; (4)~Relative Neighbor:~if the ratio between the relative distance $d_{ij}$ and the diagonal length of the whole images is less than 0.5.

\subsection{Relation Inference Module}
\label{subsec:relation}

After obtaining the whole relation embedded representation and entity embedded representation based on Eq.~(\ref{eq:relation}) and Eq.~(\ref{eq:multihead}) respectively, we can construct a global scene graph representation denoted as $\Omega(G)$:
\begin{equation}
	\begin{split}
		\Omega(G)&=\sum_{i=1}^n\Phi(f'_i), \\
		\textrm{where} \quad  f'_i&=[f_i, \sum_{j\neq i}\Theta(f_{ij})],
	\end{split}
\end{equation}
where $n$ refers to the number of entities in the image, and $\sum$ and $[\cdot]$ denote element-wise sum and concatenation operation. Then we perform recognition of entity and relation with three layers MLP as the following:
\begin{equation}
	\begin{split}
		o'_{i}&=\textrm{MLP}([f_i, \Omega(G)]), \\  l'_{ij}&=\textrm{MLP}([f_{ij}, \Omega(G)]),
	\end{split}
\end{equation}
where $o'$ and $l'$ refer to the predicted label of entity and relation, respectively. We adopt two cross-entropy loss functions in this module, and define $o$ and $l$ as the ground truth label for entity and relation, respectively:
\begin{equation}
	\label{eq:joint}
	\begin{split}
		\mathcal{L}_{entity}&=-\sum_io'_i\log(o_i), \\  \mathcal{L}_{relation}&=-\sum_i\sum_{j\neq i}l'_{ij}\log(l_{ij}).
	\end{split}
\end{equation}
In summary, the joint objective loss function in our Attentive Relational Network can be formulated as follows:
\begin{equation}
	\mathcal{L}=\lambda_1\mathcal{L}_{entity}+\lambda_2\mathcal{L}_{relation}+\lambda_3\mathcal{L}_{semantic}+\|\mathbb{W}\|^2_2,
\end{equation}
where $\lambda_1$, $\lambda_2$ and $\lambda_3$ denote hyper-parameters to tune the function, and $\mathbb{W}$ refers to all learned weights in our model.

\section{Experimental Results}
\label{sec:exp}
To validate our proposed model, extensive experiments are conducted on the public \emph{Visual Genome Dataset}~\cite{krishna2017visual}.

\subsection{Experimental Settings}

\begin{table*}[!t]
	\centering
	\caption{Comparison results of our model and existing state-of-the-art methods on constrained scene graph classification~(SGCls) and predicate classification~(PredCls) on Visual Genome~(VG)~\cite{krishna2017visual} test set. {\bf Ours w/o ST+GSA}, {\bf Ours w/ GSA}, {\bf Ours w/ ST} and {\bf Ours-Full} denote our baseline model, our model only with Graph Self-Attention Module, our model only with Semantic Transformation Module and our full model, respectively. $\dagger$ means the results obtained from corresponding papers. Results based on our implementation is marked by $\ast$. The best performances are in bold.}
	\vspace{+2mm}
	\resizebox{0.73\linewidth}{!}{
		\begin{tabular}{llcccc}
			\toprule
			\multirow{2}{*}{Dataset}  &\multirow{2}{*}{Model}  & \multicolumn{2}{c}{SGCls} & \multicolumn{2}{c}{PredCls} \\
			\cmidrule(lr){3-4}\cmidrule(lr){5-6}
			& & Recall@50  & Recall@100  & Recall@50  & Recall@100 \\ \midrule
			\multirow{18}{*}{VG} 
			& LP~\cite{lu2016visual} & 11.8 & 14.1 & 27.9 & 35.0 \\
			& Message Passing~\cite{xu2017scene} & 21.7 & 24.4 & 44.8 & 53.0 \\
			& Graph R-CNN~\cite{yang2018graph} & 29.6 & 31.6 & 54.2 & 59.1 \\
			& Neural Motif~\cite{zellers2018neural} & 35.8 & 36.5 & $55.8^\ast$/$65.2^\dagger$ & $58.3^\ast$/$67.1^\dagger$ \\
			& GPI~\cite{herzig2018mapping} & 36.5 & 38.8 & $56.3^\ast$/$65.1^\dagger$ & $60.7^\ast$/$66.9^\dagger$ \\ \cmidrule(lr){2-6}
			& ST-GSA-nosemanticloss-sum & 36.6 & 38.8 & 56.4 & 60.3 \\
			& ST-GSA-nosemanticloss-multiply & 34.0 & 36.8 & 53.5 & 59.7 \\
			& ST-GSA-nosemanticloss-concat & 36.2 & 38.4 & 55.4 & 59.9 \\ 
			& ST-GSA-sum & 36.9 & 39.1 & 56.6 & 61.1 \\
			& ST-GSA-multiply & 36.6 & 38.4 & 56.2 & 60.7 \\ 
			& ST-GSA-nowordembed & 37.3 & 39.8 & 55.7 & 60.6 \\ 
			& ST-GSA-singlehead & 37.9 & 40.1 & 56.3 & 60.9\\
			\cmidrule(lr){2-6}
			& {\bf Ours w/o ST+GSA} & 34.6 & 35.3 & 54.3 & 57.6 \\
			& {\bf Ours w/ GSA}     & 37.2 & 39.4 & 54.8 & 59.9 \\ 
			& {\bf Ours w/ ST}      & 37.3&  40.1 & 55.2 & 60.9  \\
			& {\bf Ours-Full} & {\bf 38.2} & {\bf 40.4} & {\bf 56.6} & {\bf 61.3} \\
			\cmidrule(lr){2-6}
			& {\bf Ours-Full-unconstrained} & 41.4 & 46.0 & 61.6 & 68.9 \\
			\bottomrule
		\end{tabular}%
	}
	\vspace{-2mm}
	\label{tab:result}%
\end{table*}%

{\bf Visual Genome~(VG)}~\cite{krishna2017visual} includes 108,077 images annotated with bounding boxes, entities and relationships. There are 75,729 unique object categories, and 40,480 unique relationship predicates in total. Considering the effect of long-tail distribution, we choose the most frequent 150 object categories and 50 predicates for evaluation~\cite{newell2017pixels,xu2017scene,zellers2018neural}. For a fair comparison with previous works, we follow the experimental setting in~\cite{xu2017scene}, and split the dataset into 70K/5K/32K as train/validation/test sets. 

{\bf Metrics}: Following~\cite{alexe2012measuring,lu2016visual}, we adopt the image-wise Recall$@$100 and Recall$@$50 as our evaluation metrics. Recall$@$X is used to compute the fraction of occurring times when the correct relationship is predicted in the top $x$ confident predictions. The rank strategy is based on confidence scores of objects and predicates. While, we do not choose $mAP$ as a metric, because we can not exhaustively annotate all possible relationships, and some true relationships may be missing, as discussed in~\cite{lu2016visual}. Besides, we also report per-type Recall$@$5 of classifying individual predicate.

{\bf Task Settings:} In this work, our goal is to infer the scene graph of an image given the confidence scores of entities and relations, while the object detection is not our main objective. Therefore, we conduct two sub-tasks of scene graph generation to evaluate our proposed method following~\cite{xu2017scene,herzig2018mapping}.~\textbf{(1)Scene Graph Classification~(SGCls):}~Given ground truth bounding boxes of entities, the goal is to predict the category of all entities and relations in an image. This task needs to correctly detect the triplet of $<$subject-predicate-object$>$.~\textbf{(2)Predicate Classification~(PredCls):}~Given a set of ground truth entity bounding boxes with their corresponding localization and categories, the goal is to predict all relations between entities. In all of our experiments, we perform graph-constrained evaluation, which means the returned triplets must be consistent with a scene graph. In addition, we report the results in unconstrained setting. 


{\bf Compared Methods:}~We compare our proposed approach with the following methods on VG: Language Prior~(LP)~\cite{lu2016visual}, Iterative Message Passing~(IMP)~\cite{xu2017scene}, Neural Motif~\cite{zellers2018neural}, Graph R-CNN~\cite{yang2018graph}, GPI~\cite{herzig2018mapping}. In all experiments, the parameter settings of the above-mentioned methods are adopted from the corresponding papers. Note that some of previous methods use slightly different pre-training procedures or data split or extra supervisions. For a fair comparison, we re-train Nerual Motif and GPI with their released codes for evaluation, and ensure all the methods are based on the same backbone. 


\subsection{Implementation Details}

We implement our model based on TensorFlow~\cite{girija2016tensorflow} framework on a single NVIDIA 1080 Ti GPU. Similar to prior work in scene graph generation~\cite{li2017scene,xu2017scene}, we adopt Faster R-CNN~(with ImageNet pretrained VGG16)~\cite{ren2017faster} as backbone in our object detection module. Following~\cite{li2017scene,xu2017scene,zellers2018neural}, we adopt two-stage training, where the object detection module is pre-trained for capturing label category possibility as our high-level feature. Furthermore, the semantic transformation module is implemented as three 300-size layers for semantic projection, and one fully-connected~(FC) layers for feature embedding that output a vector of size 500, and the word vectors were learned from the text data of Visual Genome with Glove~\cite{pennington2014glove}; the graph self-attention module is implemented by one FC layer that outputs a vector of size 500, and we set $k=8$ in Eq.~(\ref{eq:multihead}) as multi-head attention; the Relation Inference Module is implemented as three FC layers of size 500 and outputs an entity probability vector of size 150 and relation probability vector of size 51 corresponding to the semantic labels in the datasets. We perform an end-to-end training by employing Adam as the optimizer with initial learning rate of $1\times 10^{-4}$, and the exponential decay rate for the 1st and 2nd moment estimates are set as $0.9$ and $0.999$, respectively. We adopt a mini-batch training with batch size 20. The hyper-parameters in our joint loss function Eq.~(\ref{eq:joint}) are set as $\lambda_1:\lambda_2:\lambda_3=4:1:1$. 
\begin{table}[!t]
	\caption{Predicate classification recall of our full model on the test set of Visual Genome. Top 20 most frequent types are shown. The evaluation metric is recall$@$5.}
	\vspace{+2mm}
	\label{tab:predicate_recall}
	\centering
	\resizebox{0.7\linewidth}{!}{
		\begin{tabular}{ll||ll}
			\hline
			predicate &ours  &predicate  &ours         \\    \hline\hline
			on        &98.54 &sitting on &80.89 \\      
			has       &98.18 &between    &78.62   \\
			of        &96.17 &under      &66.17   \\ 
			wearing   &99.46 &riding     &93.01   \\
			in        &90.85 &in front of&66.29   \\
			near      &93.41 &standing on&77.84   \\
			with      &88.20 &walking on &90.05   \\
			behind    &88.72 &at         &73.19   \\
			holding   &91.44 &attached to&84.01   \\
			wears     &95.90 &belonging to&81.62   \\  \hline   
		\end{tabular}
	}
	\vspace{-5mm}
\end{table}

\subsection{Quantitative Comparisons}

As depicted in Table~\ref{tab:result}, we compare the performances of our model with the state-of-the-art methods on Visual Genome. We can see that our model outperforms all previous methods on the task of SGCls. Our full model ``Ours-Full'' achieves $38.2\%$ and $40.4\%$ w.r.t Recall$@$50 and Recall$@$100, which surpass the strong baseline method GPI by about $2\%$ in terms of both metrics. It indicates the superior capability of our model in capturing relations between entity pairs. Moreover, our full model also generates better performance in terms of PredCls, demonstrating our model's ability in recognizing relationship accurately. Noting that the PredCls task is simply trying to detect predicate that requires less structural information. While our proposed semantic transformation model and graph self-attention module perform the best in jointly learning the graph structure. Compared to other similar graph-based approach, \eg Iterative Message Passing~(IMP)~\cite{xu2017scene} and Graph R-CNN~\cite{yang2018graph}, our model can capture each node's representation by attending on the neighboring nodes to incorporate more context information and preserve the structural relationship in the image. These advantages make our model superior to~\cite{xu2017scene} and~\cite{yang2018graph}. In addition, Table~\ref{tab:predicate_recall} illustrates per-type predicate recall performances of our models on the Visual Genome test set. We find that our model achieves high Recall$@5$ of over $0.85$ in most of the frequent predicates, as well as some less frequent ones that are harder to learn, \eg `walking on' and `riding'. The reason is that our framework is able to overcome the uneven relationship distribution by better modeling contextual information and diverse graph representations. 


\subsection{Ablation Study}

In this subsection, we perform ablation studies to better examine the effectiveness of the introduced two modules. 

\begin{figure}[htbp]
	\centering
	\includegraphics[width=0.4\textwidth]{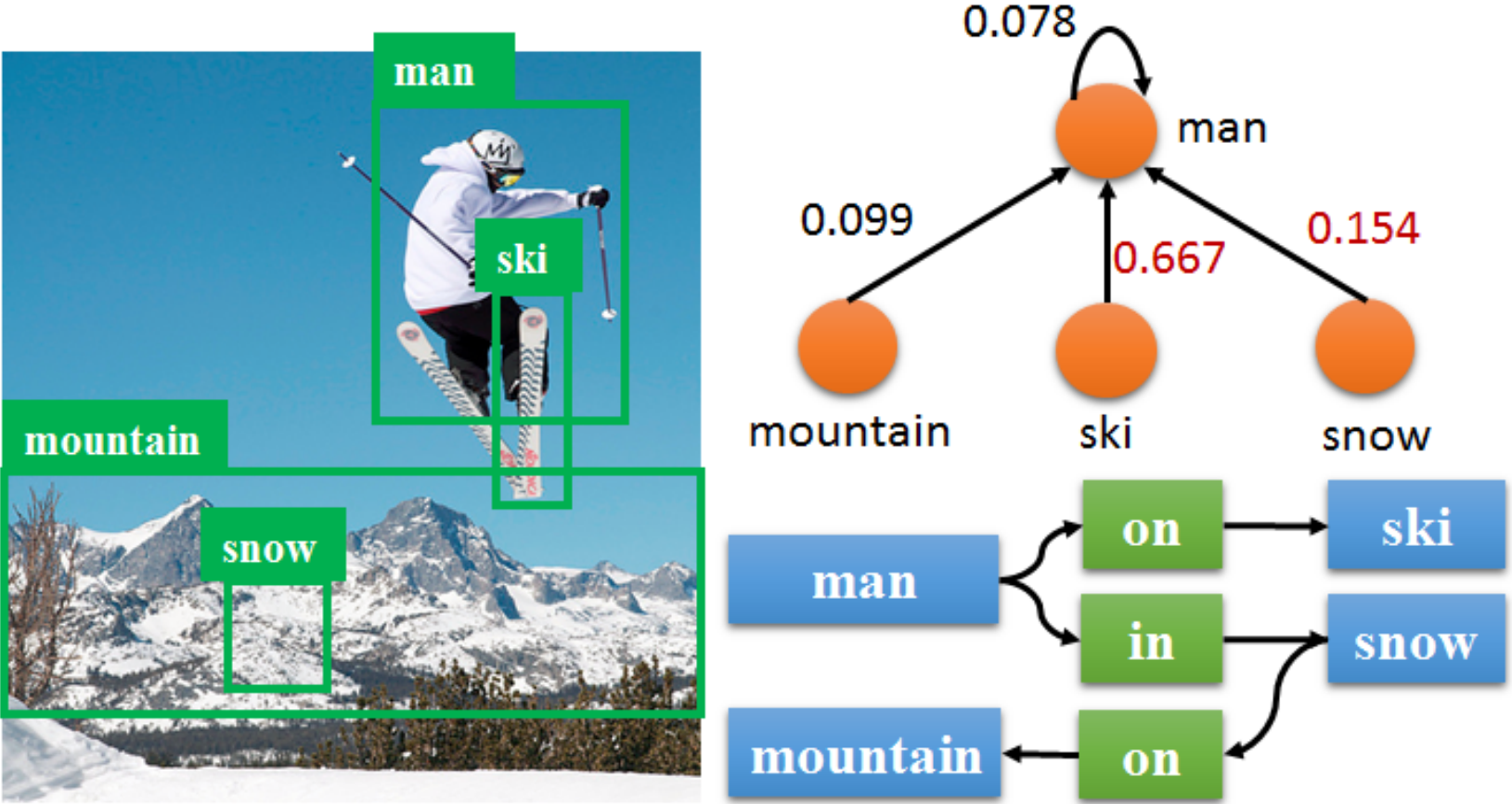}
	\caption{An example of Graph Self-Attention Module. The left illustrates the test image with object detection results. The top right shows the attention weights from other entities to the entity `man', and the bottom right depicts the ground truth scene graph.}
	\label{fig:att_example}
	\vspace{-3.5mm}
\end{figure}

\begin{figure}[htbp]
	\centering
	\includegraphics[width=0.5\textwidth]{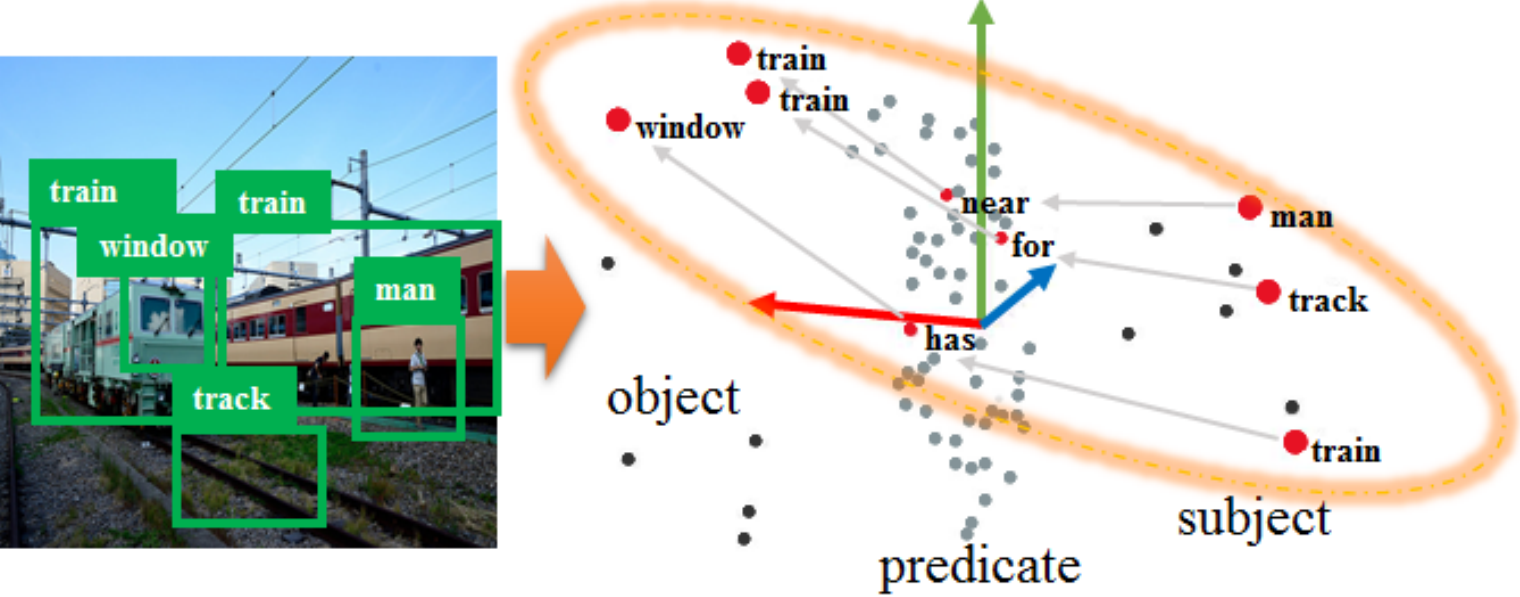}
	\vspace{-3.5mm}
	\caption{An example of Semantic Transformation Module. The left is a sample image with its entity bounding boxes visualized. The right is a PCA visualization of entity and relation features in three dimensional space on Scene Graph Classification. The red dots represent detected labels for objects, predicates and subjects.}
	\label{fig:pca}
	\vspace{-7mm}
\end{figure}

{\bf Graph Self-Attention Module:} As shown in Table~\ref{tab:result}, the graph self-attention module~(``Ours w/ GSA'') brings a large improvement compared to our baseline model~(``Ours w/o ST+GSA''). Moreover, our model with only graph self-attention module~(``Ours w/ GSA'') outperforms Neural Motif and Graph R-CNN by $2\%$ and $8\%$, respectively. The improvement is mainly brought by the attentive features generated from weighted neighbour embedding, which helps each node to focus on neighbor node features according to context relations. The overall module is thus able to capture more meaningful context across the entire graph and enhance the scene graph generation. In addition, we exploit the effectiveness of our proposed multi-head attention mechanism in the module. As shown in the middle part of Table~\ref{tab:result}, ours model with multi-head obtains slightly better performance than ours with single-head in terms of SGCls and PredCls, suggesting the multi-head can better capture useful information. Figure~\ref{fig:att_example} illustrates an example of graph self-attention helping to generate the scene graph. Our model assigns higher attention weights on `ski' to `man'~(0.667) and `snow' to `man'~(0.154) than `mountain' to `man'~(0.099), denoting the module learns to attend on more significant neighbor entities~(\eg `ski' and `snow'). The ground truth scene graph demonstrates the detected relationships match the ground truth.


\begin{figure*}[htbp]
	\centering
	\includegraphics[width=0.95\textwidth]{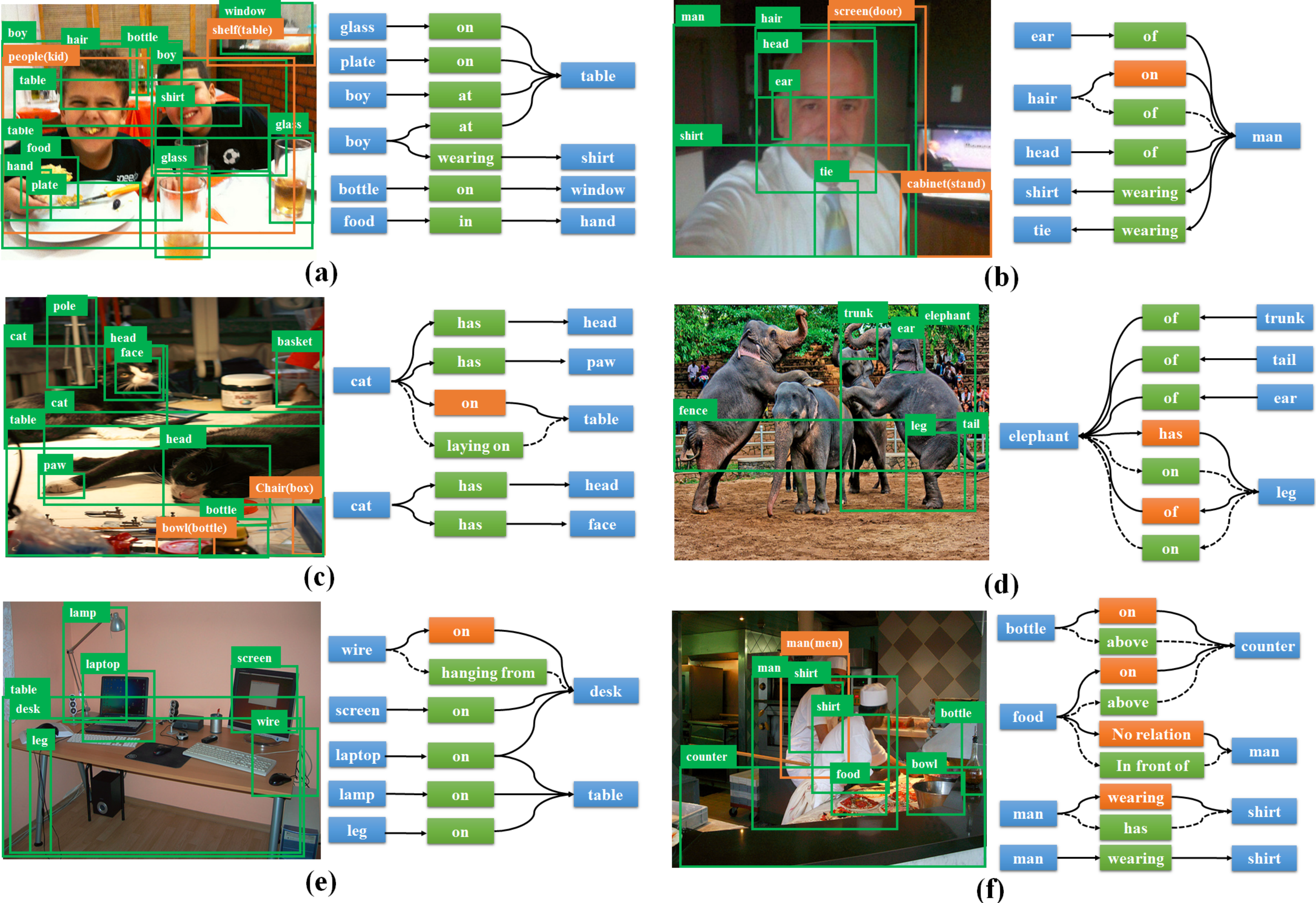}
	\caption{Qualitative results on our proposed Attentive Relational Network. Green and brown bounding boxes are correct and wrong predictions respectively~(As for brown labels, our predictions are outside the brackets, while ground truths are inside the brackets). In scene graphs, green and brown cuboids are correct and wrong relation predictions respectively. The dotted lines denote the ground truth relations mistakenly classified by our model. Only predicted boxes that overlap with the ground truth are shown.}
	\label{fig:example}
	\vspace{-3.5mm}
\end{figure*}

{\bf Semantic Transformation Module:} As shown in Table~\ref{tab:result}, our model with only semantic transformation module~(``Ours w/ ST'') outperforms all state-of-the-art results and other variants of our model, i.e. ``Ours w/o ST+GSA'' and ``Ours w/ GSA''.
This indicates the importance of the proposed semantic transformation module in generating better scene graphs. 
Furthermore, we examine the proposed semantic transformation loss function $\mathcal{L}_{semantic}$ and different approaches of feature fusion. We introduce three variants with no semantic loss for feature fusion, \ie concatenate~(``ST-GSA-nosemanticloss-concat''), sum up~(``ST-GSA-nosemanticloss-sum'') and element-wise multiply~(``ST-GSA-nosemanticloss-multiply''). Moreover, we have examined other three variants with semantic loss, \ie sum up~(``ST-GSA-sum''), element-wise multiply~(``ST-GSA-multiply''), visual feature only~(``ST-GSA-nowordembed''). As shown in Table~\ref{tab:result}, concatenating projected features through our semantic transformation achieves the best performance, suggesting our loss function, incorporating word embedding and concatenation operation is effective and necessary. By examining the PCA visualization in a 3D space illustrated in Figure~\ref{fig:pca}, we discover semantic affinities among the entity type and relation embedding of our module. Meanwhile, we notice apparent clusters of object nodes, predicate nodes and subject nodes in three dimension. Moreover, we find that the existing visual relationship can be translated into a common semantic space~(denoted as orange circle in Figure~\ref{fig:pca}), where the entity and relation nodes are connected in an approximate linearity, \eg $<$train-has-window$>$, $<$track-for-train$>$ and $<$man-near-train$>$. It demonstrates that our proposed module can learn semantic knowledge to transform visual feature and word embedding into relation space which benefits the scene graph generation tasks.

\subsection{Qualitative Results}

To qualitatively verify the constructed scene graph and visual relations learned by our proposed model, Figure~\ref{fig:example} illustrates a number of visualization examples for scene graph generation on the Visual Genome dataset. The results demonstrate that our model is able to semantically predict most of visual relations in images correctly. As an example, all of visual relationships in the scene graph are correctly detected in Figure~\ref{fig:example}~(a), which has a complex structure and several different types of objects. Moreover, our model is able to resolve the ambiguity in the object-subject direction. For instance, $<$ear-of-man$>$ and $<$man-wearing-tie$>$ are predicted correctly by our model in Figure~\ref{fig:example}~(b). In addition, we observe that our model can predict predicates more accurately than the ground truth annotations and make more reasonable correct predictions, \eg in Figure~\ref{fig:example}~(d) and (f) our model outputs $<$elephant-has-leg$>$ and $<$man-wearing-shirt$>$, while the ground truth are $<$elephant-on-leg$>$ and $<$man-has-shirt$>$ that are not inappropriate for the situation. However, there are still some failure cases in our model. First, certain mistakes stem from predicate ambiguity, \eg our model mislead in predicting $<$bottle-above-counter$>$ and $<$wire-hanging from-desk$>$ by $<$bottle-on-counter$>$ and $<$wire-on-desk$>$ in Figure~\ref{fig:example}~(f) and (e). Second, some mistakes are caused by the failure of the detector. For example, our model fails to detect any relation between `food' and `man' in Figure~\ref{fig:example}~(f), and some entities are detected inaccurately, \eg `door' and `stand' are misled by `screen' and `cabinet' in Figure~\ref{fig:example}~(b), respectively. Advanced object detection model will be beneficial for improving the performance.

\section{Conclusion}
\label{sec:con}

In this paper, we present a novel \emph{Attentive Relational Network} for scene graph generation. We introduce a semantic transformation module that projects visual features and linguistic knowledge into a common space, and a graph self-attention module for joint graph representation embedding. Extensive experiments are conducted on the \emph{Visual Genome Dataset} and our method outperforms the state-of-the-art methods on scene graph generation, which demonstrates the effectiveness of our model. \\

\noindent\footnotesize\textbf{Acknowledgement}. This work was partly supported by the National Natural Science Foundation of China (No. 61573045) and the Foundation for Innovative Research Groups through the National Natural Science Foundation of China (No. 61421003). We also would like to thank the support by NSF~(Awards No.1813709, No.1704309 and No.1722847). Mengshi Qi acknowledges the financial support from the China Scholarship Council.

{
	\bibliographystyle{ieee}
	\bibliography{scenegraph}

\begin{thebibliography}{10}\itemsep=-1pt

\bibitem{alexe2012measuring}
Bogdan Alexe, Thomas Deselaers, and Vittorio Ferrari.
\newblock Measuring the objectness of image windows.
\newblock {\em TPAMI}, 34(11):2189--2202, 2012.

\bibitem{atzmon2016learning}
Yuval Atzmon, Jonathan Berant, Vahid Kezami, Amir Globerson, and Gal Chechik.
\newblock Learning to generalize to new compositions in image understanding.
\newblock {\em arXiv preprint arXiv:1608.07639}, 2016.

\bibitem{bordes2013translating}
Antoine Bordes, Nicolas Usunier, Alberto Garcia-Duran, Jason Weston, and Oksana
  Yakhnenko.
\newblock Translating embeddings for modeling multi-relational data.
\newblock In {\em NeurlPS}, 2013.

\bibitem{dai2017detecting}
Bo Dai, Yuqi Zhang, and Dahua Lin.
\newblock Detecting visual relationships with deep relational networks.
\newblock In {\em CVPR}. IEEE, 2017.

\bibitem{farhadi2010every}
Ali Farhadi, Mohsen Hejrati, Mohammad~Amin Sadeghi, Peter Young, Cyrus
  Rashtchian, Julia Hockenmaier, and David Forsyth.
\newblock Every picture tells a story: Generating sentences from images.
\newblock In {\em ECCV}. Springer, 2010.

\bibitem{frome2013devise}
Andrea Frome, Greg~S Corrado, Jon Shlens, Samy Bengio, Jeff Dean, Tomas
  Mikolov, et~al.
\newblock Devise: A deep visual-semantic embedding model.
\newblock In {\em NeurlPS}, 2013.

\bibitem{girija2016tensorflow}
Sanjay~Surendranath Girija.
\newblock Tensorflow: Large-scale machine learning on heterogeneous distributed
  systems.
\newblock 2016.

\bibitem{girshick2015fast}
Ross Girshick.
\newblock Fast r-cnn.
\newblock In {\em ICCV}. IEEE, 2015.

\bibitem{herzig2018mapping}
Roei Herzig, Moshiko Raboh, Gal Chechik, Jonathan Berant, and Amir Globerson.
\newblock Mapping images to scene graphs with permutation-invariant structured
  prediction.
\newblock In {\em NeurlPS}, 2018.

\bibitem{hwangtensorize}
Seong~Jae Hwang, Sathya~N Ravi, Zirui Tao, Hyunwoo~J Kim, Maxwell~D Collins,
  and Vikas Singh.
\newblock Tensorize, factorize and regularize: Robust visual relationship
  learning.

\bibitem{johnson2015image}
Justin Johnson, Ranjay Krishna, Michael Stark, Li-Jia Li, David Shamma, Michael
  Bernstein, and Li Fei-Fei.
\newblock Image retrieval using scene graphs.
\newblock In {\em CVPR}. IEEE, 2015.

\bibitem{kipf2016semi}
Thomas~N Kipf and Max Welling.
\newblock Semi-supervised classification with graph convolutional networks.
\newblock In {\em ICLR}, 2017.

\bibitem{krishna2018referring}
Ranjay Krishna, Ines Chami, Michael Bernstein, and Li Fei-Fei.
\newblock Referring relationships.
\newblock In {\em CVPR}. IEEE, 2018.

\bibitem{krishna2017visual}
Ranjay Krishna, Yuke Zhu, Oliver Groth, Justin Johnson, Kenji Hata, Joshua
  Kravitz, Stephanie Chen, Yannis Kalantidis, Li-Jia Li, David~A Shamma, et~al.
\newblock Visual genome: Connecting language and vision using crowdsourced
  dense image annotations.
\newblock {\em IJCV}, 123(1):32--73, 2017.

\bibitem{lazebnik2006beyond}
Svetlana Lazebnik, Cordelia Schmid, and Jean Ponce.
\newblock Beyond bags of features: Spatial pyramid matching for recognizing
  natural scene categories.
\newblock In {\em CVPR}. IEEE, 2006.

\bibitem{li2019know}
Xiangyang Li and Shuqiang Jiang.
\newblock Know more say less: Image captioning based on scene graphs.
\newblock {\em TMM}, 2019.

\bibitem{li2017vip}
Yikang Li, Wanli Ouyang, and Xiaogang Wang.
\newblock Vip-cnn: A visual phrase reasoning convolutional neural network for
  visual relationship detection.
\newblock In {\em CVPR}. IEEE, 2017.

\bibitem{li2018factorizable}
Yikang Li, Wanli Ouyang, Bolei Zhou, Jianping Shi, Chao Zhang, and Xiaogang
  Wang.
\newblock Factorizable net: an efficient subgraph-based framework for scene
  graph generation.
\newblock In {\em ECCV}. Springer, 2018.

\bibitem{li2017scene}
Yikang Li, Wanli Ouyang, Bolei Zhou, Kun Wang, and Xiaogang Wang.
\newblock Scene graph generation from objects, phrases and region captions.
\newblock In {\em ICCV}. IEEE, 2017.

\bibitem{liang2017deep}
Xiaodan Liang, Lisa Lee, and Eric~P Xing.
\newblock Deep variation-structured reinforcement learning for visual
  relationship and attribute detection.
\newblock In {\em CVPR}. IEEE, 2017.

\bibitem{lu2016visual}
Cewu Lu, Ranjay Krishna, Michael Bernstein, and Li Fei-Fei.
\newblock Visual relationship detection with language priors.
\newblock In {\em ECCV}. Springer, 2016.

\bibitem{newell2017pixels}
Alejandro Newell and Jia Deng.
\newblock Pixels to graphs by associative embedding.
\newblock In {\em NeurlPS}, 2017.

\bibitem{newell2017associative}
Alejandro Newell, Zhiao Huang, and Jia Deng.
\newblock Associative embedding: End-to-end learning for joint detection and
  grouping.
\newblock In {\em NeurlPS}, 2017.

\bibitem{pennington2014glove}
Jeffrey Pennington, Richard Socher, and Christopher Manning.
\newblock Glove: Global vectors for word representation.
\newblock In {\em EMNLP}, 2014.

\bibitem{peyre2017weakly}
Julia Peyre, Ivan Laptev, Cordelia Schmid, and Josef Sivic.
\newblock Weakly-supervised learning of visual relations.
\newblock In {\em ICCV}. IEEE, 2017.

\bibitem{plummer2017phrase}
Bryan~A Plummer, Arun Mallya, Christopher~M Cervantes, Julia Hockenmaier, and
  Svetlana Lazebnik.
\newblock Phrase localization and visual relationship detection with
  comprehensive image-language cues.
\newblock In {\em CVPR}. IEEE, 2017.

\bibitem{qi2018stagnet}
Mengshi Qi, Jie Qin, Annan Li, Yunhong Wang, Jiebo Luo, and Luc Van~Gool.
\newblock stagnet: An attentive semantic rnn for group activity recognition.
\newblock In {\em ECCV}. Springer, 2018.

\bibitem{qi2017online}
Mengshi Qi, Yunhong Wang, and Annan Li.
\newblock Online cross-modal scene retrieval by binary representation and
  semantic graph.
\newblock In {\em MM}. ACM, 2017.

\bibitem{ramanathan2015learning}
Vignesh Ramanathan, Congcong Li, Jia Deng, Wei Han, Zhen Li, Kunlong Gu, Yang
  Song, Samy Bengio, Charles Rosenberg, and Li Fei-Fei.
\newblock Learning semantic relationships for better action retrieval in
  images.
\newblock In {\em CVPR}. IEEE, 2015.

\bibitem{redmon2016you}
Joseph Redmon, Santosh Divvala, Ross Girshick, and Ali Farhadi.
\newblock You only look once: Unified, real-time object detection.
\newblock In {\em CVPR}. IEEE, 2016.

\bibitem{ren2017faster}
Shaoqing Ren, Kaiming He, Ross Girshick, and Jian Sun.
\newblock Faster r-cnn: Towards real-time object detection with region proposal
  networks.
\newblock In {\em NeurlPS}, 2015.

\bibitem{sadeghi2011recognition}
Mohammad~Amin Sadeghi and Ali Farhadi.
\newblock Recognition using visual phrases.
\newblock In {\em CVPR}. IEEE, 2011.

\bibitem{teney2017graph}
Damien Teney, Lingqiao Liu, and Anton van~den Hengel.
\newblock Graph-structured representations for visual question answering.
\newblock In {\em CVPR}. IEEE, 2017.

\bibitem{vaswani2017attention}
Ashish Vaswani, Noam Shazeer, Niki Parmar, Jakob Uszkoreit, Llion Jones,
  Aidan~N Gomez, {\L}ukasz Kaiser, and Illia Polosukhin.
\newblock Attention is all you need.
\newblock In {\em NeurlPS}, 2017.

\bibitem{velickovic2017graph}
Petar Velickovic, Guillem Cucurull, Arantxa Casanova, Adriana Romero, Pietro
  Lio, and Yoshua Bengio.
\newblock Graph attention networks.
\newblock In {\em ICLR}, 2018.

\bibitem{woo2018linknet}
Sanghyun Woo, Dahun Kim, Donghyeon Cho, and In~So Kweon.
\newblock Linknet: Relational embedding for scene graph.
\newblock In {\em NeurlPS}, 2018.

\bibitem{xu2017scene}
Danfei Xu, Yuke Zhu, Christopher~B Choy, and Li Fei-Fei.
\newblock Scene graph generation by iterative message passing.
\newblock In {\em CVPR}. IEEE, 2017.

\bibitem{yang2018graph}
Jianwei Yang, Jiasen Lu, Stefan Lee, Dhruv Batra, and Devi Parikh.
\newblock Graph r-cnn for scene graph generation.
\newblock In {\em ECCV}. Springer, 2018.

\bibitem{yang2018shuffle}
Xu Yang, Hanwang Zhang, and Jianfei Cai.
\newblock Shuffle-then-assemble: learning object-agnostic visual relationship
  features.
\newblock In {\em ECCV}. Springer, 2018.

\bibitem{yao2018exploring}
Ting Yao, Yingwei Pan, Yehao Li, and Tao Mei.
\newblock Exploring visual relationship for image captioning.
\newblock In {\em ECCV}. Springer, 2018.

\bibitem{yin2018zoom}
Guojun Yin, Lu Sheng, Bin Liu, Nenghai Yu, Xiaogang Wang, Jing Shao, and
  Chen~Change Loy.
\newblock Zoom-net: Mining deep feature interactions for visual relationship
  recognition.
\newblock In {\em ECCV}. Springer, 2018.

\bibitem{yu2017visual}
Ruichi Yu, Ang Li, Vlad~I Morariu, and Larry~S Davis.
\newblock Visual relationship detection with internal and external linguistic
  knowledge distillation.
\newblock In {\em ICCV}. IEEE, 2017.

\bibitem{yuan2017temporal}
Yuan Yuan, Xiaodan Liang, Xiaolong Wang, Dit-Yan Yeung, and Abhinav Gupta.
\newblock Temporal dynamic graph lstm for action-driven video object detection.
\newblock In {\em ICCV}. IEEE, 2017.

\bibitem{zellers2018neural}
Rowan Zellers, Mark Yatskar, Sam Thomson, and Yejin Choi.
\newblock Neural motifs: Scene graph parsing with global context.
\newblock In {\em CVPR}. IEEE, 2018.

\bibitem{zhang2017visual}
Hanwang Zhang, Zawlin Kyaw, Shih-Fu Chang, and Tat-Seng Chua.
\newblock Visual translation embedding network for visual relation detection.
\newblock In {\em CVPR}. IEEE, 2017.

\bibitem{zhang2017ppr}
Hanwang Zhang, Zawlin Kyaw, Jinyang Yu, and Shih-Fu Chang.
\newblock Ppr-fcn: weakly supervised visual relation detection via parallel
  pairwise r-fcn.
\newblock In {\em ICCV}. IEEE, 2017.

\bibitem{zhang2018grounding}
Hanwang Zhang, Yulei Niu, and Shih-Fu Chang.
\newblock Grounding referring expressions in images by variational context.
\newblock In {\em CVPR}. IEEE, 2018.

\bibitem{zhang2017relationship}
Ji Zhang, Mohamed Elhoseiny, Scott Cohen, Walter Chang, and Ahmed Elgammal.
\newblock Relationship proposal networks.
\newblock In {\em CVPR}. IEEE, 2017.

\bibitem{zhou2014learning}
Bolei Zhou, Agata Lapedriza, Jianxiong Xiao, Antonio Torralba, and Aude Oliva.
\newblock Learning deep features for scene recognition using places database.
\newblock In {\em NeurlPS}, 2014.

\bibitem{zhuang2017towards}
Bohan Zhuang, Lingqiao Liu, Chunhua Shen, and Ian Reid.
\newblock Towards context-aware interaction recognition for visual relationship
  detection.
\newblock In {\em ICCV}. IEEE, 2017.

\end{thebibliography}
}

\end{document}